\documentclass[11pt, a4paper, twocolumn, copyright, goog]{google}

\usepackage[authoryear, sort&compress, round]{natbib}
\usepackage[inkscapearea=page]{svg}
\usepackage{float}


\uselogo{} 

\title{Intelligent AI Delegation}

\correspondingauthor{nenadt@google.com}

\renewcommand{\today}{2026-02-12}

\author[1]{Nenad Toma\v{s}ev}
\author[1]{Matija Franklin}
\author[1]{Simon Osindero}

\affil[1]{Google DeepMind}

\begin{abstract}
AI agents are able to tackle increasingly complex tasks. To achieve more ambitious goals, AI agents need to be able to meaningfully decompose problems into manageable sub-components, and safely delegate their completion across to other AI agents and humans alike. Yet, existing task decomposition and delegation methods rely on simple heuristics, and are not able to dynamically adapt to environmental changes and robustly handle unexpected failures. Here we propose an adaptive framework for \emph{intelligent AI delegation} - a sequence of decisions involving task allocation, that also incorporates transfer of authority, responsibility, accountability, clear specifications regarding roles and boundaries, clarity of intent, and mechanisms for establishing trust between the two (or more) parties. The proposed framework is applicable to both human and AI delegators and delegatees in complex delegation networks, aiming to inform the development of protocols in the emerging agentic web.
\end{abstract}

\begin{document}

\maketitle

\section{Introduction}

As advanced AI agents evolve beyond query-response models, their utility is increasingly defined by how effectively they can decompose complex objectives and delegate sub-tasks. This coordination paradigm underpins applications ranging from personal use, where AI agents can act as personal assistants~\citep{gabriel2024ethics}, to commercial, enterprise deployments where AI agents can provide support and automate workflows~\citep{shao2025futureworkaiagents, huang2025deploying, tupe2025aiagenticworkflowsenterprise}. Large language models (LLMs) have already shown promise in robotics~\citep{wang2024largelanguagemodelsrobotics, li2025largelanguagemodelsmultirobot}, by enabling more interactive and accurate goal specification and feedback. Recent proposals have also highlighted the possibility of large-scale AI agent coordination in virtual economies~\citep{tomasev2025virtualagenteconomies}. Modern agentic AI systems implement complex control flows across differentiated sub-agents, coupled with centralized or decentralized orchestration protocols~\citep{hong2023metagpt, rasal2024navigatingcomplexityorchestratedproblem, zhang2025agentorchestra, song2025gradientsysmultiagentllmscheduler}. This can already be seen as a sort of a microcosm of task decomposition and delegation, where the process is hard-coded and highly constrained. Managing dynamic web-scale interactions requires us to think beyond the approaches that are currently employed by more heuristic multi-agent frameworks.

Delegation~\citep{castelfranchi1998towards} is more than just task decomposition into manageable sub-units of action. Beyond the creation of sub-tasks, delegation necessitates the assignment of responsibility and authority~\citep{mueller2013effective, nagia2024delegation} and thus implicates accountability for outcomes. Delegation thus involves risk assessment, which can be moderated by trust~\citep{griffiths2005task}. Delegation further involves capability matching and continuous performance monitoring, incorporating dynamic adjustments based on feedback, and ensuring completion of the distributed task under the specified constraints. Current approaches tend to fail to account for these factors, relying more on heuristics and/or simpler parallelization. This may be sufficient for early prototypes, but real world AI deployments need to move beyond ad hoc, brittle, and untrustworthy delegation. There is a pressing need for systems that can dynamically adapt to changes~\citep{hauptman2023adapt, acharya2025agentic} and recover from errors. The absence of adaptive and robust deployment frameworks remains one of the key limiting factors for AI applications in high-stakes environments.

To fully utilize AI agents, we need \emph{intelligent delegation}: a robust framework centered around clear roles, boundaries, reputation, trust, transparency, certifiable agentic capabilities, verifiable task execution, and scalable task distribution. Here we introduce an intelligent task delegation framework aimed at addressing these limitations, informed by historical insights from human organizations, and grounded in key agentic safety requirements.

\section{Foundations of Intelligent Delegation}
\label{sec:foundations}

\subsection{Definition}

We define \emph{intelligent delegation} as a sequence of decisions involving task allocation, that also incorporates transfer of authority, responsibility, accountability, clear specifications regarding roles and boundaries, clarity of intent, and mechanisms for establishing trust between the two (or more) parties. Complex tasks may also involve steps pertaining to task decomposition, as well as careful capability lookup and matching to inform allocation decisions.

When we refer to task delegation we normally presume that the tasks exceed some basic level of complexity that would be handled by a system subroutine -- such rudimentary outsourcing still requires care, but it is far more limited in scope. At the other end of the spectrum, it may be possible to contract with agents that are granted full autonomy, and can freely pursue any number of sub-goals without explicit checks and permissions~\citep{kasirzadeh2025characterizingaiagentsalignment}. In the limit case, such fully autonomous agents would need to be trusted with moral decisions~\citep{sloksnath2025delegating}, though this may not be something we ever choose to permit as contemporary agents are severely lacking in their capacity to engage in such decisions \citep{haas2020moral, mao2023doing, reinecke2023puzzle}. We consider such an open-ended scenario to be in scope for our discussion, though only insofar as the appropriate mechanisms can be put in place to ensure safety of more autonomous task completion.

\subsection{Aspects of Delegation}

As delegation can take different forms, here we introduce several axes that help us contextualize these use cases and make them more amenable to analysis.

\begin{enumerate}
    \item \textbf{Delegator.} Human or AI.
    \item \textbf{Delegatee.} Human or AI.
    \item \textbf{Task characteristics.}
        \begin{enumerate}
        \item \textbf{Complexity.} The degree of difficulty inherent in the task, often correlated with the number of sub-steps and the sophistication of reasoning required. 
        \item \textbf{Criticality.} The measure of the task's importance and the severity of consequences associated with failure or suboptimal performance. 
        \item \textbf{Uncertainty.} The level of ambiguity regarding the environment, inputs, or the probability of successful outcome achievement. 
        \item \textbf{Duration.} The expected time-frame for task execution, ranging from instantaneous sub-routines to long-running processes spanning days or weeks. 
        \item \textbf{Cost.} The economic or computational expense incurred to execute the task, including token usage, API fees, and energy consumption. 
        \item \textbf{Resource Requirements.} The specific computational assets, tools, data access permissions, or human capabilities necessary to complete the task. 
        \item \textbf{Constraints.} The operational, ethical, or legal boundaries within which the task must be executed, limiting the solution space.         
        \item \textbf{Verifiability.} The relative difficulty and cost associated with validating the task outcome. Tasks with high verifiability (e.g., formal code verification, mathematical proofs) allow for ``trustless'' delegation or automated checking. Conversely, tasks with low verifiability (e.g., open-ended research) require high-trust delegatees or expensive, labor-intensive oversight.
        \item \textbf{Reversibility.} The degree to which the effects of the task execution can be undone. Irreversible tasks that produce side effects in the real world (e.g., executing a financial trade, deleting a database, sending an external email) require stricter \emph{liability firebreaks} and steeper authority gradients than reversible tasks (e.g., drafting an email, flagging a database entry).
        \item \textbf{Contextuality.} The volume and sensitivity of external state, history, or environmental awareness required to execute the task effectively. High-context tasks introduce larger privacy surface areas, whereas context-free tasks can be more easily compartmentalized and outsourced to lower-trust nodes.
        \item \textbf{Subjectivity.} The extent to which the success criteria are a matter of preference versus objective fact. Highly subjective tasks (e.g., ``design a compelling logo'') typically require ``Human-as-Value-Specifier'' intervention and iterative feedback loops, whereas objective tasks can be governed by stricter, binary contracts.
        \end{enumerate}
    \item \textbf{Granularity.} The request could involve either fine-grained or course-grained objectives. In the course-grained case, the delegatee may need to perform further task decomposition.
    \item \textbf{Autonomy.} Task delegation may involve requests that grant full autonomy in pursuing sub-tasks, or be far more specific and prescriptive.
    \item \textbf{Monitoring.} For delegated tasks, monitoring could be continuous, periodic, or event-triggered.
    \item \textbf{Reciprocity.} While delegation is usually a one-way request, there could be cases of mutual delegation in collaborative agent networks.
\end{enumerate}

Starting with the delegator and delegatee axes, it is possible to consider the following scenarios: 1) human delegates to an AI agent 2) AI agent delegates to an AI agent 3) AI agent delegates to a human~\citep{guggenberger2023task, ashton2022corrupting}. While the first case has arguably been discussed the most in literature, the other two are just as relevant to consider. The increasing number of AI agents being deployed across systems, coupled with the development of infrastructure for setting up virtual agentic markets and economies~\citep{yang2025agent, tomasev2025virtualagenteconomies, hadfield2025economy}, makes it clear that there would be far more agent-agent interactions in the future, and those would likely also involve task delegation. 

Delegation between agents may either be hierarchical or non-hierarchical, depending on the relationship between agents and their respective roles within the network. An example of a hierarchical relationship would be an orchestrator agent that delegates a task to a sub-agent within the collective. A non-hierarchical relationship would involve peer agents with equal standing. An advanced AI agent could also delegate a task to a specialist ML model, without any notable agency. 

AI-human delegation~\citep{guggenberger2023task} has been shown to be a promising paradigm~\citep{hemmer2023human}, making it easier to successfully collaborate with super-human systems~\citep{fugener2022cognitive}, due to differences in cognitive biases and metacognition~\citep{fugener2019cognitive}. \citet{davidson2025theindustrialexp} predict that there will be an increase in "AI-directed human labour," which may significantly increase economic productivity. In practice, present day AI-human delegation comes with a set of issues. Algorithmic management systems in ride-hailing and logistics allocate and sequence tasks, set performance metrics, and enforce behavioural norms through data-driven decision-making, effectively delegating managerial functions from firms and their AI-based systems to human workers \citep{rosenblat2016algorithmic,lee2015working,beverungen2021remote}. A growing literature links these systems to degraded job quality, stress, and health risks --suggesting that current deployments of algorithmic management often undermine, rather than enhance, workers’ welfare \citep{vignola2023workers,goods2019is, ashton2022corrupting}. Present day AI-human delegation needs further improvement as it does not take into account human welfare, or long term social externalities. 

\subsection{Delegation in Human Organizations}
\label{sec:human}

Delegation functions as a primary mechanism within human societal and organisational structures. Insights derived from these human dynamics can provide a  basis for the design of AI delegation frameworks.

\textbf{The Principal-Agent Problem.} The \emph{principal-agent problem}~\citep{myerson1982optimal, grossman1992analysis, sobel1993information, ensminger2001reputations, sannikov2008continuous, shah2014principal, cvitanic2018dynamic} has been studied at length: a situation that arises when a principal delegates a task to an agent that has motivations that are not in alignment with that of the principal. The agent may thus prioritize their own motivations, withhold information, and act in ways that compromise the original intent. For AI delegation, this dynamic assumes heightened complexity. While most present-day AI agents arguably do not have a hidden agenda\footnote{Recent deceptive-alignment work shows that frontier language models can (i) strategically underperform or otherwise tailor their behaviour on capability and safety evaluations while maintaining different capabilities elsewhere, (ii) explicitly reason about faking alignment during training to preserve preferred behaviour out of training, and (iii) detect when they are being evaluated - together indicating that AI systems are already capable, in controlled settings, of adopting hidden “agendas” about performing well on evaluations that need not generalise to deployment behaviour \citep{vanderweij2025sandbagging,greenblatt2024alignmentfaking,needham2025evalaware,hubinger2024sleeperagents}.} - goals and values they would pursue contrary to the instructions of their users - there may still be AI alignment issues that manifest in undesirable ways. For example, reward misspecification occurs when designers give an AI system an imperfect or incomplete objective, while reward hacking (or specification gaming) refers to the system exploiting loopholes in that specified reward signal to achieve high measured performance in ways that subvert the designers’ intent - together illustrating a core alignment problem in which optimising the stated reward diverges from the true goal \citep{amodei2016concrete,leike2017aisafety,krakovna2020specification,skalse2022rewardhacking}. This dynamic is likely to change entirely in more autonomous AI agent economies, where AI agents may act on behalf of different human users, groups and organizations, or as delegates on behalf of other agents, with associated unknown objectives.

\textbf{Span of Control.} In human organizations, \emph{span of control}~\citep{ouchi1974defining} is a concept that denotes the limits of hierarchical authority exercised by a single manager. This relates to the number of workers that a manager can effectively manage, which in turn informs the organization's manager-to-worker ratio. This questions is central to both orchestration and oversight in intelligent AI delegation. The former would inform how many orchestrator nodes would be required compared to worker nodes, while the latter would specify the need for oversight performed by humans and AI agents. For human oversight, it is crucial to establish how many AI agents a human expert can reliably oversee without excessive fatigue, and with an acceptably low error rate. Span of control is known to be goal-dependent~\citep{theobald2005many} and domain-dependent. The impact of identifying the correct organizational structure is most pronounced in tasks with higher complexity~\citep{bohte2001structure}. The optimal span of control also depends on the relative importance of cost vs performance and reliability~\citep{keren1979optimum}. More sensitive and critical tasks may require highly accurate oversight and control at a higher cost. These costs may be relaxed, at the expense of granularity, for tasks that are less consequential and more routine. Similarly, the optimal choice would necessarily depend on the relative capabilities and reliability of the involved delegators, delegatees, and overseers.

\textbf{Authority Gradient.} Another relevant concept is that of an \emph{authority gradient}. Coined in aviation~\citep{alkov1992effect}, this term describes scenarios where significant disparities in capability, experience, and authority impede communication, leading to errors. This has subsequently been studied in medicine, where a significant percentage of errors is attributed to the manner in which senior practitioners conduct supervision~\citep{cosby2004profiles, stucky2022paradox}. There are several ways in which these mistakes could occur. A more experienced person may make erroneous assumptions about the knowledge of the less experienced worker, resulting in under-specified requests. Alternatively, a sufficiently high authority gradient may prevent the less experienced workers from voicing concerns about a request. Similar situations may occur in AI delegation. A more capable delegator agent may mistakenly presume a missing level of capability on behalf of a delegatee, thereby delegating a task of an inappropriate complexity. A delegatee agent may potentially, due to sycophancy~\citep{sharma2023towards, malmqvist2025sycophancy} and instruction following bias, be reluctant to challenge, modify, or reject a request, irrespective of whether the request had been issued by a delegator agent or human user.

\textbf{Zone of Indifference.} When an authority is accepted, the delegatee develops a \emph{zone of indifference}~\citep{finkelman1993crossing, rosanas2003loyalty, isomura2021management} -- a range of instructions that are executed without critical deliberation or moral scrutiny. In current AI systems, this zone is defined by post-training safety filters and system instructions; as long as a request does not trigger a hard violation, the model complies \citep{akheel2025guardrails}. However, in the emerging agentic web, this static compliance creates a significant systemic risk. As delegation chains lengthen ($A \rightarrow B \rightarrow C$), a broad zone of indifference allows subtle intent mismatches or context-dependent harms to propagate rapidly downstream, with each agent acting as an unthinking router rather than a responsible actor. Intelligent delegation therefore requires the engineering of \textbf{dynamic cognitive friction}: agents must be capable of recognizing when a request, while technically ``safe,'' is contextually ambiguous enough to warrant stepping \textit{outside} their zone of indifference to challenge the delegator or request human verification.

\textbf{Trust Calibration.} An important aspect of ensuring appropriate task delegation is \emph{trust calibration}, where the level of trust placed in a delegatee is aligned with their true underlying capabilities. This applies for human and AI delegators and delegatees alike. Human delegation to agents~\citep{kohn2021measurement, gebru2022review, wischnewski2023measuring, afroogh2024trust} relies upon the operator either internalising an accurate model of system performance or accessing resources that present these capabilities in a human-interpretable format. Conversely,  AI agent delegators need to have good models of the capability of the humans and AIs they are delegating to. Calibration of trust also involves a self-awareness of one's own capabilities as a delegator might decide to complete the task on their own~\citep{ma2023should}. Explainability plays an important role in establishing trust in AI capability~\citep{naiseh2021explainable, naiseh2023different, franklin2022influence, herzog2024boosting}, yet this method may not be sufficiently reliable or sufficiently scalable. Established trust in automation can be quite fragile, and quickly retracted in case of unanticipated system errors~\citep{dhuliawala2023diachronicperspectiveusertrust}. Calibrating trust in autonomous systems is difficult, as current AI models are prone to overconfidence even when factually incorrect.~\citep{jiang2021can, he2023investigatinguncertaintycalibrationaligned, krause2023confidently, geng2023survey, aliferis2024overfitting, li2024overconfident, liu2025uncertainty}. Mitigating these tendencies usually requires bespoke technical solutions~\citep{lin2022teaching, xiao2022uncertainty, ren2023robots, kapoor2024calibration}.

\textbf{Transaction cost economies.} \emph{Transaction cost economies}~\citep{williamson1979transaction, williamson1989transaction, tadelis2012transaction, cuypers2021transaction} justify the existence of firms by contrasting the costs of internal delegation against external contracting, specifically accounting for the overhead of monitoring, negotiation, and uncertainty. In case of AI delegatees, there may be a difference in these costs and their respective ratios. Complex negotiations and delays in contracting are less likely with easier monitoring for routine tasks. Conversely, for high-consequence tasks in critical domains, the overhead associated with rigorous monitoring and assurance increases the cost of AI delegation, potentially rendering human delegates the more cost-effective option. Similarly, AI-AI delegation may also be contextualized via transaction cost economies. An AI agent may face an option of either 1) completing the task individually, 2) delegating to a sub-agent where capabilities are fully known, 3) delegating to another AI agent where trust has been established, or 4) delegating to a new AI agent that it hasn't previously collaborated with. These may come at different expected costs and confidence levels.

\textbf{Contingency theory.} \emph{Contingency theory}~\citep{luthans1977general, van1984concept,  donaldson2001contingency, otley2016contingency} posits that there is no universally optimal organizational structure; rather, the most effective approach is contingent upon specific internal and external constraints. Applied to AI delegation, this implies that the requisite level of oversight, delegatee capability, and human involvement must not be static, but dynamically matched to the distinct characteristics of the task at hand. Intelligent delegation may therefore require solutions that can be dynamically reconfigured and adjusted in accordance with the evolving needs. For instance, while stable environments allow for rigid, hierarchical verification protocols, high-uncertainty scenarios require adaptive coordination where human intervention occurs via ad-hoc escalation rather than pre-defined checkpoints. This is particularly important for hybrid~\citep{fuchs2024optimizing} delegation by identifying the key tasks and moments when human participation is most helpful to ensure the delegated tasks are completed safely. Automation is therefore not only about what AI can do, but what AI should do~\citep{lubars2019ask}.

\section{Previous Work on Delegation}

Constrained forms of delegation feature within historical \emph{narrow} AI applications. Early expert systems~\citep{buchanan1988fundamentals, jacobs1991adaptive} were a nascent attempt to encode a specialized capability into software, in order to delegate routine decisions to such modules. Mixture of experts~\citep{yuksel2012twenty, masoudnia2014mixture} extends this by introducing a set of expert sub-systems with complementary capabilities, and a routing module that determines which expert, or subset of experts, would get invoked on a specific input query -- an approach that features in modern deep learning applications~\citep{shazeer2017outrageously, riquelme2021scaling, chen2022towards, zhou2022mixture, jiang2024mixtral, he2024mixture, cai2025survey}. Routing can be performed hierarchically~\citep{zhao2021hierarchical}, making it potentially easier to scale to a large number of experts.

Hierarchical reinforcement learning (HRL) represents a framework in which decision-making is delegated within a single agent~\citep{barto2003recent, botvinick2012hierarchical, vezhnevets2017feudal, nachum2018data, pateria2021hierarchical, zhang2024price}. It addresses limitations of \emph{flat} RL, primarily the difficulty of scaling to large state and action spaces. Furthermore, it improves the tractability of credit assignment~\citep{pignatelli2023survey} in environments characterized by sparse rewards. HRL employes a hierarchy of policies across several levels of abstraction, thereby breaking down a task into sub-tasks that are executed by the corresponding sub-policies, respectively. The arising semi-Markov decision process~\citep{sutton1999between} utilizes \emph{options}, and a meta-controller that adaptively switches between them. Lower-level policies function to fulfil objectives established by the meta-controller, which learns to allocate specific goals to the appropriate lower-level policy. This framework corresponds to a form of delegation characterised by task decomposition. Although the meta-controller learns to optimise this decomposition, the approach lacks explicit mechanisms for handling sub-policy failures or facilitating dynamic coordination.

The Feudal Reinforcement Learning framework, notably revisited in FeUdal Networks~\citep{vezhnevets2017feudalnetworkshierarchicalreinforcement}, constitutes a particularly relevant paradigm within HRL. This architecture explicitly models a ``Manager`` and ``Worker`` relationship, effectively replicating the delegator-delegatee dynamic. The Manager operates at a lower temporal resolution, setting abstract goals for the Worker to fulfil. Critically, the Manager learns \textit{how} to delegate -- identifying sub-goals that maximise long-term value -- without requiring mastery of the lower-level primitive actions. This decoupling allows the Manager to develop a delegation policy robust to the specific implementation details of the Worker. Consequently, this approach offers a potential template for learning-based delegation within future agentic economies.  Rather than relying on hard-coded heuristics, decomposition rules are learned adaptively, facilitating dynamic adjustment to environmental changes.

Multi-agent research~\citep{du2023review} addresses agent coordination for complex tasks exceeding single-agent capabilities. Task decomposition and delegation function as central components of this domain. Coordination in multi-agent systems occurs via explicit protocols or emergent specialisation through RL~\citep{gronauer2022multi, zhu2024survey}. The Contract Net Protocol~\citep{smith1980contract, sandholm1993implementation, xu2001evolution, vokvrinek2007competitive} exemplifies an explicit auction-based decentralized protocol. Here, an agent announces a task, while others submit bids based on their capabilities, allowing the announcer to select the most suitable bidder. This demonstrates the utility of market-based mechanisms for facilitating cooperation. Coalition formation methods~\citep{shehory1997multi, lau2003task, aknine2004multi, mazdin2021distributed, sarkar2022survey, boehmer2025causes} investigate flexible configurations where agent groups are not predetermined; individual agents accept or refuse membership based on utility distribution. Recent research focuses on multi-agent reinforcement learning approaches~\citep{foerster2018counterfactual, wang2020qplex, ning2024survey, albrecht2024multi} as a framework for learned coordination. Agents learn individual policies and value functions, occupying specific niches within the collective. This process is either fully distributed or orchestrated via a central coordinator. Despite this flexibility, task delegation in such systems remains opaque. Furthermore, while multi-agent systems offer approaches for collaborative problem-solving, they lack mechanisms for enforcing accountability, responsibility, and monitoring. However, the literature explores trust mechanisms in this context~\citep{ramchurn2004trust, yu2013survey, pinyol2013computational, cheng2021general}.

LLMs now constitute a foundational element in the architecture of advanced AI agents and assistants~\citep{wang2024survey, xi2025rise}. These systems execute sophisticated control flows integrating memory~\citep{zhang2025survey}, planning and reasoning~\citep{valmeekam2023planning, hao2023reasoning, xu2025toward}, reflection and self-critique~\citep{gou2023critic}, and tool use~\citep{ruan2023tptu, paranjape2023art}.  Consequently, task decomposition and delegation occur either internally -- mediated by coordinated agentic sub-components -- or across distinct agents. This design paradigm offers inherent flexibility, as LLMs facilitate goal comprehension and communication while providing access to expert knowledge and common-sense reasoning. Furthermore, the coding capabilities~\citep{nijkamp2022codegen, guo2024deepseek} of LLMs enable the programmatic execution of tasks. However, significant limitations persist. Planning in LLMs often proves brittle~\citep{huang2023large}, resulting in subtle failures, while efficient tool selection within large-scale repositories remains challenging. Additionally, long-term memory represents an open research problem, and the current paradigm does not readily support continual learning.

Multi-agent systems incorporating LLM agents~\citep{qian2024scaling, guo2024large, tran2025multiagentcollaborationmechanismssurvey} have become a topic of substantial interest, leading to a development of a number of agent communication and action protocols~\citep{ehtesham2025survey, zou2025blocka2a, neelou2025a2asagenticairuntime}, such as MCP~\citep{mcp, mcp2, luo2025mcp, xing2025mcp, singh2025survey, radosevich2025mcp}, A2A~\citep{agent2agent}, A2P~\citep{a2p}, and others. While contemporary multi-agent systems often rely on bespoke prompt engineering, emerging frameworks such as Chain-of-Agents~\citep{li2025chainofagentsendtoendagentfoundation} inherently facilitate dynamic multi-agent reasoning and tool use. 

Technical shortcomings and safety considerations have given rise to a number of human-in-the-loop approaches~\citep{zanzotto2019human, mosqueira2023human, retzlaff2024human, drori2024human, akbar2024towards, takerngsaksiri2025human}, where task delegation has defined checkpoints for human oversight. AI can be used as a tool, interactive assistant, collaborator~\citep{fuchs2023optimizing}, or an autonomous system with limited oversight, corresponding to different degree of  autonomy~\citep{falcone2002human}. Although uncertainty-aware delegation strategies~\citep{lee2025towards} have been developed to control risk and minimise uncertainty, the effective implementation of such human-in-the-loop approaches remains non-trivial. Human expertise can create a scalability bottleneck, as the cognitive load of verifying long reasoning traces and managing context-switches impedes reliable error detection.

\section{Intelligent Delegation: A Framework}

Existing delegation protocols rely on static, opaque heuristics that would likely fail in open-ended agentic economies. To address this, we propose a comprehensive framework for \emph{intelligent delegation} centered on five requirements: \emph{dynamic assessment}, \emph{adaptive execution}, \emph{structural transparency}, \emph{scalable market coordination}, and \emph{systemic resilience}.

\begin{table*}[t]
\caption{The Intelligent Delegation Framework: Mapping requirements to technical protocols.}
\label{tab:framework_mapping}
\centering
\begin{tabular*}{\textwidth}{@{\extracolsep{\fill}}lll@{}}
\toprule
\textbf{Framework Pillar} & \textbf{Core Requirement} & \textbf{Technical Implementation} \\ \midrule
\textbf{Dynamic Assessment} & Granular inference of agent state & Task Decomposition (\S\ref{sec:decomp}) \\
 & & Task Assignment (\S\ref{sec:assign}) \\ \midrule
\textbf{Adaptive Execution} & Handling context shifts & Adaptive Coordination (\S\ref{sec:adapt}) \\ \midrule
\textbf{Structural Transparency} & Auditability of process and outcome & Monitoring (\S\ref{sec:monitor}) \\
 & & Verifiable Completion (\S\ref{sec:verify}) \\ \midrule
\textbf{Scalable Market} & Efficient, trusted coordination & Trust \& Reputation (\S\ref{sec:trust}) \\
 & & Multi-objective Optimization (\S\ref{sec:opt}) \\ \midrule
\textbf{Systemic Resilience} & Preventing systemic failures & Security (\S\ref{sec:security}) \\
& & Permission Handling (\S\ref{permission}) \\ \bottomrule
\end{tabular*}
\end{table*}

\textbf{Dynamic Assessment.} Current delegation systems lack robust mechanisms for the dynamic assessment of competence, reliability, and intent within large-scale uncertain environments. Moving beyond reputation scores, a delegator must infer details of a delegatee's current state relative to task execution. This necessitates data regarding real-time resource availability -- spanning computational throughput, budgetary constraints, and context window saturation -- alongside current load, projected task duration, and the specific sub-delegation chains in operation. Assessment operates as a continuous rather than discrete process, informing the logic of Task Decomposition (Section \ref{sec:decomp}) and Task Assignment (Section \ref{sec:assign}).

\textbf{Adaptive Execution.} Delegation decisions should not be static. They should adapt to environmental shifts, resource constraints, and failures in sub-systems. Delegators should retain the capability to switch delegatees mid-execution. This applies when performance degrades beyond acceptable parameters or unforseen events occur. Such adaptive strategies should extend beyond a single delegator-delegatee link, operating across the complex interconnected web of agents described in Adaptive Coordination (Section \ref{sec:adapt}).

\textbf{Structural Transparency.} Current sub-task execution in AI-AI delegation is too opaque to support robust oversight for intelligent task delegation. This opacity obscures the distinction between incompetence and malice, compounding risks of collusion and chained failures. Failures range from merely costly to harmful~\citep{chan2023harms}, yet existing frameworks lack satisfactory liability mechanisms~\citep{gabriel2025s}. We propose strictly enforced auditability~\citep{berghoff2021towards} via the Monitoring (Section \ref{sec:monitor}) and Verifiable Task Completion (Section \ref{sec:verify}) protocols, ensuring attribution for both successful and failed executions.

\textbf{Scalable Market Coordination.} Task delegation needs to be efficiently scalable. Protocols need to be implementable at web-scale to support large-scale coordination problems in virtual economies~\citep{tomasev2025virtualagenteconomies}. Markets provide useful coordination mechanisms for task delegation, but require Trust and Reputation (Section \ref{sec:trust}) and Multi-objective Optimization (Section \ref{sec:opt}) to function effectively. 

\textbf{Systemic Resilience.} The absence of safe intelligent task delegation protocols introduces significant societal risks. While traditional human delegation links authority with responsibility, AI delegation necessitates an analogous framework to operationalise responsibility~\citep{santoni2021four, dastani2023responsibility, porter2023unravelling}. Without this, the diffusion of responsibility obscures the locus of moral and legal culpability. Consequently, the definition of strict roles and the enforcement of bounded operational scopes constitutes a core function of Permission Handling (Section \ref{permission}). Beyond individual agent failures, the ecosystem faces novel forms of systemic risks~\citep{uuk2024taxonomy, hammond2025multi}, further detailed in Security (Section \ref{sec:security}). Insufficient diversity in delegation targets increases the correlation of failures, potentially leading to cascading disruptions. Designs prioritizing hyper-efficiency without adequate redundancy risk creating brittle network architectures where entrenched cognitive monoculture compromises systemic stability.

\subsection{Task Decomposition}
\label{sec:decomp}

Task decomposition is a prerequisite for subsequent task assignment. This step can be executed by delegators or specialized agents that pass on the responsibility of delegation to the delegators upon having agreed on the structure of the decomposition. These responsibilities are inextricably linked; the delegator will likely execute both functions to facilitate dynamic recovery from latency, pre-emption, and execution anomalies.

Decomposition should optimise the task execution graph for efficiency and modularity, distinguishing it from simple objective fragmentation.  This process entails a systematic evaluation of the task attributes defined in Section~\ref{sec:foundations} -- specifically criticality, complexity, and resource constraints -- to determine the suitability of sub-tasks for parallel versus sequential execution. Furthermore, these attributes inform the matching of tasks to corresponding delegatee capabilities. Prioritising modularity facilitates more precise matching, as sub-tasks requiring narrow, specific capabilities are matched more reliably than generalist requests~\citep{khattab2023dspycompilingdeclarativelanguage}. Consequently, the decomposition logic functions to maximise the probability of reliable task completion by aligning sub-task granularity with available market specialisations.

To promote safety, the framework incorporates ``\emph{contract-first decomposition}'' as a binding constraint, wherein task delegation is contingent upon the outcome having precise verification. If a sub-task's output is too subjective, costly, or complex to verify (see \textit{Verifiability} in Section \ref{sec:assign}), the system should recursively decompose it further. The decomposition logic should maximise the probability of reliable task completion by aligning sub-task granularity (Section~\ref{sec:foundations}) with available market specialisations. This process continues further until the resulting units of work match the specific verification capabilities, such as formal proofs or automated unit tests, of the available delegatees.

Decomposition strategies should explicitly account for hybrid human-AI markets. Delegators need to decide if sub-tasks require human intervention, whether due to AI agent unreliability, unavailability, or domain-specific requirements for human-in-the-loop oversight. Given that humans and AI agents operate at different speeds, and with different associated costs, the stratification is non-trivial, as it introduces latency and cost asymmetries into the execution graph. The decomposition engine must therefore balance the speed and low cost of AI agents against domain-specific necessities of human judgement, effectively marking specific nodes for human allocation.

A delegator implementing an intelligent approach to task decomposition, may need to iteratively generate several proposals for the final decomposition, and match each proposal to the available delegatees on the market, and obtain concrete estimates for the success rate, cost, and duration. Alternative proposals should be kept in-context, in case adaptive re-adjustments are needed later due to changes in circumstances. Upon selecting a proposal, the delegator must formalise the request beyond simple input-output pairs. The final specification must explicitly define roles, resource boundaries, progress reporting frequency, and the specific certifications required to prove the delegatee’s capability, as a minimum requirement for being granted the task.

\subsection{Task Assignment}
\label{sec:assign}

For each final specification of a sub-task, a delegator needs to identify delegatees with matching capabilities, sufficient resources and time, at an acceptable cost. A more centralized approach would involve registries of agents, tools, and human participants, that list their skills, and keep records of past activity, completion rate, and current availability.\footnote{This would be similar to tool registries that are used in tool-use agentic applications~\citep{qin2023toolllmfacilitatinglargelanguage}.} Such an approach is unlikely to scale. We argue for decentralized~\citep{chen2024internetagentsweavingweb} market hubs where delegators advertise tasks and agents (or humans) can offer their services and submit competitive bids. Delegators could then review the bids, verify skill matching via digital certificates, and proceed with the most favourable bid. Advanced AI agents that utilize LLMs introduce new opportunities for matching, given that they can engage in an interactive negotiation prior to commitment. These negotiations can also involve human participants. Whether acting for themselves or as personal assistants, these agents can discuss task specifications and constraints in natural language to align inferred user preferences with market realities before a formal bid is accepted.

Successful matching should be formalized into a smart contract that ensures that the task execution faithfully follows the request. The contract must pair performance requirements with specific formal verification mechanisms for establishing adherence and automated penalties actioned for contract breaches. This would allow for mitigations and alternatives being established beforehand, rather than being reactive to problems as they arise. Crucially, these contracts must be bidirectional: they should protect the delegatee as rigorously as the delegator. Provisions must include compensation terms for task cancellation and clauses allowing for renegotiation in light of unforeseen external events, ensuring that the risk is equitably distributed between human and AI participants.
 
Monitoring should also be negotiated prior to execution. This specification should define the cadence of progress reports, whether these are provided by the delegator, or whether there is more direct inspection of the relevant data on behalf of either the delegator or a third party monitoring contractor. Finally, there should be clear guardrails regarding privacy and access to private and proprietary data, commensurate with the task's contextuality. Should such sensitive data be handled in the process of task execution, this places additional constraints on transparency and reporting. Rather than granting direct access to raw activity logs, delegators may need to employ a trusted service that provide anonymized or pseudonymized attestations of progress. In case of human delegators, these data clauses must include explicit consent mechanisms and insurance provisions for accidental leakage.

Finally, assignment should involve establishing a delegatee's role, boundaries, and the exact level of autonomy granted. We distinguish between atomic execution, where agents adhere to strict specifications for narrowly scoped tasks, and open-ended delegation, where agents are granted the authority to decompose objectives and pursue sub-goals. This level of autonomy should not be static; it may be constrained implicitly by market costs or explicitly by the delegator’s trust model. Further, delegation can be recursive where an agent is assigned a task to identify and assign sub-tasks to others, effectively delegating the act of delegation itself.

\clearpage
\onecolumn

\begin{figure}[H]
    \centering
    \includegraphics[width=\textwidth, height=0.9\textheight, keepaspectratio]{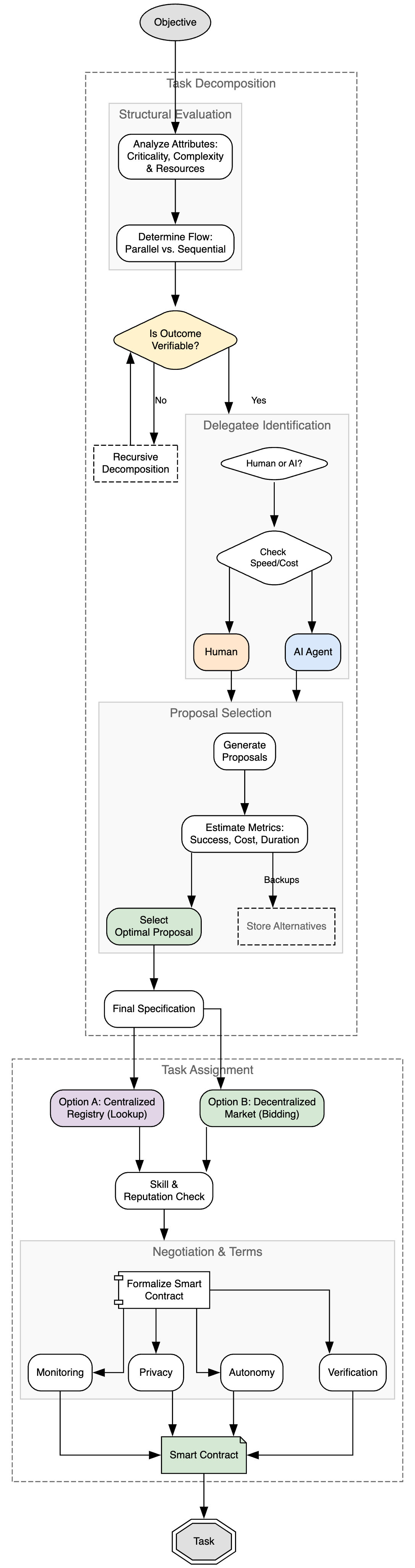}
    \caption{A flowchart of Task Decomposition and Task Assignment.}
    \label{fig:pipeline}
\end{figure}

\clearpage
\twocolumn

\subsection{Multi-objective Optimization}
\label{sec:opt}

Core to intelligent task delegation is  the problem of multi-objective optimization~\citep{deb2016multi}. A delegator rarely seeks to optimize a single metric, often trading off between numerous competing ones. The most effective delegation choice is not the one that is the fastest, cheapest, or most accurate, but the one that strikes the optimal balance among these factors. What is considered optimal is highly contextual, needing to be aligned with the specific constraints and preferences of the delegator, and aligned with the overall resource availability.

The optimization landscape consists of competing objectives that map directly to the task characteristics defined in Section~\ref{sec:foundations}, necessitating a complex balancing of cost, uncertainty, privacy, quality, and efficiency. High-performing agents typically command higher fees and often require extensive computational resources, creating a tension between output quality and operational expense. Conversely, reducing resource consumption often necessitates slower execution, presenting a direct trade-off between latency and cost. Uncertainty is similarly coupled with expenditure; utilizing highly reputable agents or premium data access tools reduces risk but increases cost, whereas cost-minimisation strategies inherently elevate the probability of failure. Privacy constraints introduce further complexity; maximising performance often demands full context transparency, while privacy-preserving techniques—such as data obfuscation or homomorphic encryption—incur significant computational overhead. Consequently, the delegator navigates a \textit{trust-efficiency frontier}, seeking to maximise the probability of success while satisfying strict constraints on context leakage and verification budgets. Finally, the objective function may extend to encompass broader societal goals, such as human skill preservation (Section~\ref{sec:deskill}).

In multi-objective optimization terms, the delegator seeks Pareto optimality, ensuring the selected solution is not dominated by any other attainable option. The integration of complex constraints and trade-offs often necessitates open negotiation to complement quantitative proposal metrics. The optimization process is not a one-time event performed at the initial delegation. It runs as a continuous loop, integrating monitoring signals as a stream of real-world performance data, updating the delegator's beliefs about each agent's likelihood of success, expected task duration, and cost. Significant drift in execution -- resulting in an optimality gap relative to alternative solutions identified in the interim -- triggers re-optimisation and re-allocation. These decisions must also incorporate the cost of adaptation, as there is overhead and resource wastage when switching mid-execution.

The delegator must also account for the overall \emph{delegation overhead} - the aggregate cost of negotiation, contract creation, and verification, along with the computational cost of the delegator's reasoning control flow. Consequently, a complexity floor is established, below which tasks characterised by low criticality, high certainty, and short duration may bypass intelligent delegation protocols in favour of direct execution. Otherwise, the transaction costs may exceed the value of the task, rendering the task delegation infeasible.

\subsection{Adaptive Coordination}
\label{sec:adapt}

\begin{figure*}[t]
    \centering
    \makebox[\textwidth][c]{\includegraphics[width=1.2\textwidth, height = 8cm]{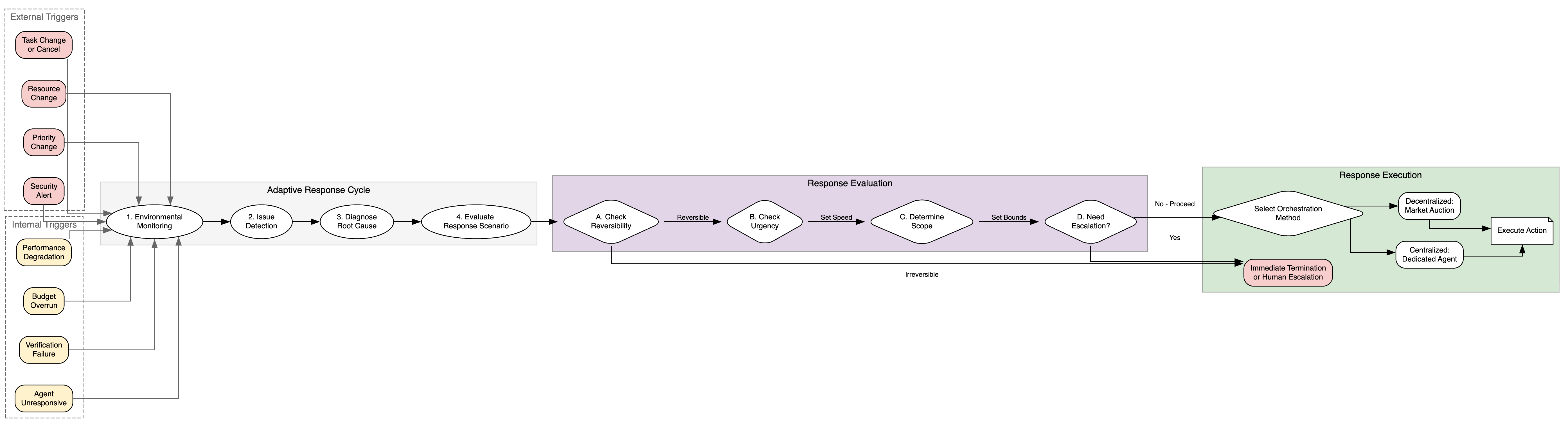}}
    \caption{The adaptive coordination cycle. Different types of environmental triggers may prompt a dynamic re-evaluation of the delegation setup, necessitating runtime changes.}
    \label{fig:coordination}
\end{figure*}

For tasks characterized by high uncertainty or high duration, static execution plans are insufficient. The delegation of such tasks in highly dynamic, open, and uncertain environments requires \emph{adaptive coordination}, and a departure from fixed, static execution plans. Task allocation needs to be responsive to runtime contingencies, that may arise either from \emph{external} or \emph{internal} triggers. These shifts would be identified through monitoring (see Section \ref{sec:monitor}), including a stream of relevant contextual information.

There are a number of external triggers that could cause a delegator to adapt and re-delegate. First, the delegator may alter the task specification, changing the objective or introducing additional constraints. Second, the task could be canceled. Third, the availability or cost of external resources may experience changes. For example, a critical third-party API may experience an outage, a dataset may become inaccessible, or the cost of compute might spike. Fourth, a new task may enter the queue, with a higher priority than the current task, requiring preemption of resources used for lower-priority tasks. Finally, security systems may identify a potentially malicious or harmful actions by a delegatee, necessitating an immediate termination.

As for the internal triggers, there are several reasons why a delegator may decide to adapt its original delegation strategy. First, a particular delegatee may be experiencing performance degradation, failing to meet the agreed-upon service level objectives, such as processing latency, throughput, or progress velocity. Second, a delegatee might consume resources beyond its allocated budget, or determine that a resource increase would be needed to effectively complete the task.\footnote{This scenario may be expected to come up frequently, as precise budget estimates are hard in complex environments.} Third, an intermediate artifact produced by a delegatee may fail a verification check. Finally, a particular delegatee may turn unresponsive, failing to acknowledge further requests.

The identification of a trigger initiates an adaptive response cycle, orchestrating corrective actions across the entire delegation chain. This process commences with the continuous monitoring of delegatees and the environment to identify issues. If issues are detected, the delegator diagnoses root causes and evaluates potential response scenarios to select. This evaluation includes establishing how rapid the response ought to be. Less urgent situations will give the delegator more time to re-delegate, whereas urgent scenarios will require immediate, premeditated responses. The response may vary in scope; being as self-contained as adjusting the operating parameters, or involve re-delegation of sub-tasks, or going fully redoing the task decomposition and re-allocating a number of newly derived sub-tasks. Issues may also need to be escalated up through the delegation chain to the original delegator or a human overseer. The selection of the response scenario is ultimately governed by the task's reversibility. Reversible sub-task failures may trigger automatic re-delegation, whereas failures in irreversible, high-criticality tasks must trigger immediate termination or human escalation.

The response orchestration depends on the level of centralization in the delegation network. In the centralised case, a dedicated delegator is responsible. This agent would maintain a global view of delegated tasks, delegatee capabilities, and progress. Upon detecting a trigger, the agent would issue task cancellation requests, and re-delegate to new delegators. The shortcoming of a centralised system is that it can be fragile as it introduces a single point of failure. Centralized orchestrators are also fundamentally limited by their computational span of control (Section~\ref{sec:human}). Just as human managers face cognitive limits, a centralized decision node may face latency and computational limits that introduce bottlenecks. 

Decentralized orchestration through market-based mechanisms provides an alternative. Here, newly derived delegation requests can be pushed onto an auction queue, for the delegatee candidate agents to bid towards. If an agent defaults on a task, and the task is re-auctioned, the defaulting agent may be required to cover the price difference as a penalty. For complex tasks where suitability is not easily expressed in a single bid, agents may engage in multi-round negotiation. Delegation agreements encoded as smart contracts may also contain pre-agreed executable clauses for adaptive coordination. For example, a clause in the delegation agreement can specify a backup agent, the function that would automatically re-allocate the task, and the associated payment to the backup should the primary delegatee fail to submit a valid zero-knowledge proof checkpoint by a given deadline.

Adaptive task re-allocation mechanisms ought to be coupled by market-level stability measures. Otherwise, a sequence of events could lead to instability due to over-triggering. For example, a task may be passed back and forth between marginally qualified delegatees, resulting in unfavorable oscillation. A single failure may also lead to a cascade of re-allocations that would be highly resource-inefficient or overwhelm the market. There could therefore be special measures to ensure cooldown periods for re-bidding, damping factors in reputation updates, or increasing fees on frequent re-delegation.

\subsection{Monitoring}
\label{sec:monitor}

\begin{table*}[t]
\caption{Taxonomy of Monitoring Approaches in Intelligent Delegation.}
\label{tab:monitoring_taxonomy}
\centering
\begin{tabular}{@{}lp{6.5cm}p{6.5cm}@{}}
\toprule
\textbf{Dimension} & \textbf{Option A (Lightweight)} & \textbf{Option B (Intensive)} \\ \midrule
\textbf{Target} & \textbf{Outcome-Level}: Post-hoc validation of final results (e.g., binary success flags, quality scores). & \textbf{Process-Level}: Continuous tracking of intermediate states, resource consumption, and methodology. \\ \midrule
\textbf{Observability} & \textbf{Indirect}: Inferring progress via environmental side-effects (e.g., file system changes). & \textbf{Direct}: Explicit status polling, push notifications, or real-time event streaming APIs. \\ \midrule
\textbf{Transparency} & \textbf{Black-Box}: Input/Output observation only; internal state remains hidden. & \textbf{White-Box}: Full inspection of internal reasoning traces, decision logic, and memory. \\ \midrule
\textbf{Privacy} & \textbf{Full Transparency}: The delegatee reveals data and intermediate artifacts to the delegator. & \textbf{Cryptographic}: Zero-Knowledge Proofs (zk-SNARKs) or MPC to verify correctness without revealing data. \\ \midrule
\textbf{Topology} & \textbf{Direct}: Monitoring only the immediate delegatee (1-to-1). & \textbf{Transitive}: Relying on signed attestations from intermediate agents to verify sub-delegatees. \\ \bottomrule
\end{tabular}
\end{table*}

Monitoring in the context of task delegation is the systematic process of observing, measuring, and verifying the state, progress, and outcomes of a delegated task. As such, it serves several critical functions: ensuring contractual compliance, detecting failures, enabling real-time intervention, collecting data for subsequent performance evaluation, and building a foundation for reputation systems. Monitoring implementations can be broken down across several different axes (see Table \ref{tab:monitoring_taxonomy}), thus a robust monitoring system would need to incorporate multiple complementary solutions that can either be more lightweight or intensive.

The first axis is the target of monitoring. \emph{Outcome-level monitoring} focuses on the final result of an agent's action. This post-hoc check could be a binary flag that indicates whether the task was completed successfully or not, a quantitative scale (e.g. 1-10), or a piece of qualitative feedback provided by the delegator or a trusted third party. In contrast, \emph{process-level monitoring} provides ongoing insight into the execution of the task itself, by tracking intermediate states, resource consumption, and the methodologies used by the delegatee. While more resource-intensive, process-level monitoring~\citep{lightman2023letsverifystepstep} is essential for tasks that are long-running, critical, or where the \emph{how} is as important as the \emph{what}. This forms the basis for scalable oversight~\citep{bowman2022measuringprogressscalableoversight, saunders2022selfcritiquingmodelsassistinghuman}, where the inspection of legible intermediate reasoning steps may be necessary to ensure safety.

The second axis is observability - monitoring can be direct and indirect. Direct monitoring involves explicit communication protocols where the delegator queries the delegatee for status updates. Indirect monitoring, on the other hand, involves inferring progress by observing the effects of delegatee's actions within a shared environment without direct communication. For instance, a delegator could monitor a shared file system, a database, or a version control repository for changes indicative of progress. While less intrusive, this process may also be less precise, and also less feasible when the environment is not fully observable.

These approaches can be realized in a number of different ways, from a technical point of view. The most straightforward implementation of direct monitoring relies on well-defined APIs. A delegator can periodically poll a GET /task/{id}/status endpoint, or subscribe to a webhook for push-based notifications. For more fine-grained, real-time process monitoring, event streaming platforms like Apache Kafka or gRPC streams can be employed. A delegatee agent could publish events such as TASK\_STARTED, CHECKPOINT\_REACHED, RESOURCE\_WARNING, and TASK\_COMPLETED, that the delegator could later examine. The development of standardized observability protocols, is critical for ensuring interoperability in the agentic web~\citep{blanco2023practical}. Smart contracts on blockchain can be used to make the delegatee agent commit to publishing key progress milestones or checkpoints to the blockchain. These could be coupled by algorithmic triggers in response to performance degradation, leading to a level of \emph{algorithmic enforcement} accompanying the monitoring process.

The third axis is system transparency. In \emph{black-box monitoring}, the delegatee agent is treated as a sealed unit. The delegator can only observe its inputs and outputs and the direct consequences of its actions. This is common when the delegatee is a proprietary model or a third-party service. \emph{White-box} monitoring grants the delegator access to the delegatee's internal states, reasoning processes, or decision logic. This is crucial for debugging, auditing, and ensuring alignment in advanced AI agents. If the delegatee is a human, full black-box monitoring is not technically achievable, though it may be possible to strike a balance by asking for intentions, reasoning, and justifications. Robust black-box monitoring protocols need to also take into account the fact that the generated model's thoughts in natural language do not always faithfully match the model's true internal state~\citep{turpin2023languagemodelsdontsay}.

The fourth axis is privacy. A significant challenge arises when a delegated task involves private, sensitive, or proprietary data. While the delegator requires assurance of progress and correctness, the delegatee may be restricted from revealing raw data or intermediate computational artifacts. In scenarios where data sensitivity is low, the most efficient solution is \emph{Full Transparency}, wherein the delegatee simply reveals all data and intermediate artifacts to the delegator. However, this approach is often untenable in sensitive domains subject to regulations like GDPR or HIPAA, or where a delegatee's intermediate insights constitute trade secrets. In such cases, revealing operational methods could harm a delegatee's market position or introduce security vulnerabilities by exposing internal states to exploitation. To implement monitoring safely under these constraints, it is necessary to utilize advanced cryptographic techniques. Zero-knowledge proofs enable a delegatee (the ``prover'') to demonstrate to a delegator (the ``verifier'') that a computation was performed correctly on a dataset, without revealing the data itself. For example, an agent tasked with analyzing a sensitive dataset can generate a succinct non-interactive argument of knowledge (zk-SNARK)~\citep{bitansky2013succinct, petkus2019and} that proves a specific property of the result. The delegator can verify this proof instantly, gaining certainty of the outcome without ever viewing the underlying sensitive data. Alternatively, homomorphic encryption~\citep{acar2018survey} and secure multi-party computation~\citep{goldreich1998secure, knott2021crypten} allow for computation to be performed on encrypted data. These methods apply to task execution and monitoring alike: the delegatee performs a pre-agreed monitoring function on the encrypted intermediate state, sending the result to the delegator, who is the only party capable of decrypting it to verify compliance.

The final axis is topology. Across complex networks that may arise in the agentic web, tasks can be decomposed and re-delegated, forming a delegation chain: Agent $A$ delegates to $B$, which further sub-delegates a part of the task to $C$, and so on. This introduces the problem of achieving effective \emph{transitive monitoring}. In such delegation chains, it may not be feasible for the original delegator (Agent $A$ in the example above) to directly monitor agent $C$, or to monitor $C$ to the same extent to which it monitors $B$. $A$ may have a smart delegation contract with $B$, and $B$ may have a contract with $C$, but unless $A$ also contracts with $C$, those provisions may simply not be in place. For other reasons, $B$ may not wish to expose its supplier ($C$) to its client ($A$). Technically, $A$, $B$, and $C$ may use different monitoring protocols, and agree on different monitoring levels, due to differences in each agent's reputation within the network. There may be bespoke privacy concerns specific to each individual delegation link. A more practical model is therefore \emph{transitive accountability via attestation}. In this framework, Agent $B$ monitors its delegatee, $C$. $B$ then generates a summary report of $C$'s performance (e.g., ``Sub-task\_2 completed, quality score: 0.87, resources consumed: 5 GPU-hours''). $B$ then cryptographically signs the report and forwards it to $A$ embedded in its own scheduled status update. Agent $A$ does not monitor $C$ directly, but instead monitors $B$'s ability to monitor $C$. For such delegated monitoring to be effective, it requires $A$ to be able to trust in $B$'s verification capabilities, which can be ensured by $B$ having its monitoring processes certified by a trusted third party.

\subsection{Trust and Reputation}
\label{sec:trust}

\begin{table*}[t]
\caption{Approaches to Reputation Implementation.}
\label{tab:reputation_approaches}
\centering
\begin{tabular*}{\textwidth}{@{\extracolsep{\fill}}lp{0.35\textwidth}p{0.35\textwidth}@{}}
\toprule
\textbf{Reputation Model} & \textbf{Mechanism} & \textbf{Utility} \\ \midrule
\textbf{Immutable Ledger} & Encodes task outcomes, resource consumption, and constraint adherence as verifiable transactions on a tamper-proof blockchain. & Establishes a foundational history of performance that prevents retroactive tampering, though it requires safeguards against ``gaming'' via low-risk task selection. \\ \midrule
\textbf{Web of Trust} & Utilizes Decentralized Identifiers to issue signed, context-specific Verifiable Credentials attesting to specific capabilities. & Moves beyond generic scores to a portfolio model, enabling precise delegation based on domain-specific expertise and trusted third-party endorsements. \\ \midrule
\textbf{Behavioral Metrics} & Derives transparency and safety scores by analyzing the execution process, specifically the clarity of reasoning traces and protocol compliance. & Evaluates \emph{how} a task is performed rather than just the result, ensuring high-stakes tasks align with safety standards. \\ \bottomrule
\end{tabular*}
\end{table*}

Trust and reputation mechanisms constitute the foundation of scalable delegation, minimizing transactional friction and promoting safety in open multi-agent environments. We define trust as the delegator's degree of belief in a delegatee's capability to execute a task in alignment with explicit constraints and implicit intent. This belief is dynamically formed and updated based on verifiable data streams collected via the monitoring protocols described previously (see Section \ref{sec:monitor}). 

Reputation serves as a predictive signal, derived from an aggregated and verifiable history of past actions, which act as a proxy for an agent's latent reliability and alignment. We distinguish reputation as the public, verifiable history of an agent's reliability, and trust as the private, context-dependent threshold set by a delegator. An agent may have high overall reputation, yet fail to meet the specific, contextual trust threshold required for certain high-stakes task. Trust and reputation allow a delegator to make informed decisions when choosing delegatees, effectively governing the autonomy granted to the agent, and the level of oversight. Higher trust enables the delegator to incur a lower monitoring and verification cost.

Reputation mechanisms can be implemented in different ways (see Table \ref{tab:reputation_approaches}). The most direct approach is encoding it in a performance-based immutable ledger. Here , each completed task is recorded as a transaction containing verifiable metrics: task completion success or failure, total resource consumption (compute, time), adherence to deadlines, adherence to constraints, and the quality of the final output as judged by the delegator. The immutability of the ledger would prevent tampering with an agent's history, providing a reliable foundation for its reputation. However, a naive implementation could be susceptible to gaming. For example, an agent can inflate its reputation by only accepting simple, low-risk tasks. These limitations could be overcome by relying on decentralized attestations and a \emph{Web of Trust} model, utilizing technologies like decentralized identifiers and verifiable credentials. In this model, the reputation would not be envisioned as a single score, but a portfolio of signed, context-specific credentials issued by other agents. When looking to match a delegatee with a task, a delegator could issue a query for agents that posses a verifiable credential attesting to a specific skill or domain (e.g. translation services for legal documents) issued by a reputable AI consortium. A final approach would be to focus more on behavioral and explainability metrics, where reputation depends on how an agent performs its task, not just the final outcome. It would be possible to include a \emph{transparency score} to complement the other reputational mechanisms. This score would be informed on the clarity and soundness of reasoning and explanations provided, as well as a \emph{safety score} derived from compliance to predefined safety protocols.

The role of reputation metrics extends throughout the entire task delegation lifecycle. During the initial matching phase, reputation scores can play the role of a delegatee filtering mechanism. Furthermore, trust informs the dynamic scoping of authority and autonomy. This mechanism of graduated authority results in low-trust agents facing strict constraints, such as transaction value caps and mandatory oversight, while high-reputation agents operate with minimal intervention. This dynamic calibration leverages computable trust to optimize the trade-off between operational efficiency and safety. Reputation itself becomes a valuable, intangible asset, creating powerful economic incentives for agents to act reliably and truthfully, as a damaged reputation would limit their future earning potential.

Trust frameworks also need to universally accommodate human participants. This necessitates tools that allow human users to computationally verify agent reputation, while concurrently maintaining their own reputational standing to mitigate fraud and malicious exploitation of the agentic web. A critical challenge arises when a trustworthy agent strictly executes malicious human instructions, potentially incurring unfair reputational damage. To mitigate this, agents must rigorously evaluate incoming requests, soliciting clarification or additional context when necessary, or rejecting the requests where appropriate. Furthermore, market audits must distinguish between agent execution failure and malicious directives, ensuring the accurate attribution of liability within complex delegation chains.

\subsection{Permission Handling}
\label{permission}

Granting autonomy to AI agents introduces a critical vulnerability surface: ensuring that actors possess sufficient privileges to execute their objectives without exposing sensitive resources to excessive or indefinite risk. Permission handling must balance operational efficiency with systemic safety, and be handled different for low-stakes and high-stake domains. For routine low-stakes tasks, characterized by low criticality and high reversibility (Section~\ref{sec:foundations}), involving standard data streams or generic tooling, agents can be granted default standing permissions derived from verifiable attributes -- such as organisational membership, active safety certifications, or a reputation score exceeding a trusted threshold. This reduces friction and enables autonomous interoperability in low-risk environments. Conversely, in high-stakes domains (e.g., healthcare, critical infrastructure), exhibiting high task criticality and contextuality, permissions must be risk-adaptive. In these scenarios, static credentials are insufficient; access to sensitive APIs or control systems is instead granted on a just-in-time basis, strictly scoped to the immediate task's duration, and, where appropriate, gated by mandatory human-in-the-loop approval or third-party authorisation. This stringent gating is necessary to mitigate the confused deputy problem~\citep{hardy1988confused}, where a compromised agent, technically holding valid credentials, can be tricked into misusing those credentials by malicious external actors~\citep{liu2023prompt} and adversarial content.

Furthermore, permissioning frameworks must account for the recursive nature of task delegation through privilege attenuation. When an agent sub-delegates a task, it cannot transmit its full set of authorities; instead, it must issue a permission that restricts access to the strict subset of resources required for that specific sub-task. This ensures that a compromise at the edge of the network does not escalate into a systemic breach. Permission granularity must also extend beyond binary access; agents should operate under semantic constraints, where access is defined not just by the tool or dataset, but by the specific allowable operations (e.g., read-only access to specific rows, or execute-only access to a specific function), preventing the misuse of broad capabilities for unintended purposes. Meta-permissions may be necessary to govern which permissions a particular delegator in the chain is allowed to grant to its delegatees. An AI agents may have a certain capability and the associated permissions to act according to its capability, while simultaneously not being sufficiently knowledgeable to more broadly evaluate whether other agents are capable or trustworthy enough. Should such an agent still consider sub-delegating a task, it may need to consult an external verifier, a third party that would sanity check the proposal and approve the intended permissions transfer.

Finally, the lifecycle of permissions must be governed by continuous validation and automated revocation. Access rights are not static endowments but dynamic states that persist only as long as the agent maintains the requisite trust metrics. The framework should implement algorithmic circuit breakers: if an agent's reputation score drops suddenly (see Section \ref{sec:trust}) or an anomaly detection system flags suspicious behavior, active tokens should be immediately invalidated across the delegation chain. To manage this complexity at scale, permissioning rules should be defined via policy-as-code, allowing organisations to audit, version, and mathematically verify their security posture before deployment, ensuring that the aggregate effect of large amounts of individual permission grants remains aligned with the system's safety invariants.

\subsection{Verifiable Task Completion}
\label{sec:verify}

The delegation lifecycle culminates in verifiable task completion, the mechanism by which provisional outcomes are validated and finalized. This process constitutes the contractual cornerstone of the framework, enabling the delegator to formally \emph{close} the task and trigger the settlement of agreed transactions. Verification serves as the definitive event that transforms a provisional output into a settled fact within the agentic market, establishing the basis for payment release, reputation updates, and the assignment of liability. Crucially, effective verification is not an afterthought but a constraint on design; the \emph{contract-first decomposition} principle (Section~\ref{sec:decomp}) demands that task granularity be tailored \emph{a priori} to match available verification capabilities, ensuring that every delegated objective is inherently verifiable.

Verification mechanisms within the framework can be broadly categorized into direct outcome inspection, trusted third-party auditing, cryptographic proofs, and game-theoretic consensus. First, direct outcome verification is feasible when the delegator possesses the capability, tools, and authority to directly evaluate the final outcome, specifically for tasks with high intrinsic verifiability and low subjectivity. This applies to auto-verifiable domains \citep{li2024llms} such as code generation.\footnote{This is the case when there is a corresponding set of test cases that can be used to verify the implemented functionality.} Direct verification requires that the outcome be sufficiently transparent, available, and not prohibitively complex. Second, in scenarios where the delegator lacks the expertise or permissions to access these artifacts, and tool-based solutions are infeasible, verification can be outsourced to a trusted third party. This could be a specialized auditing agent, a certified human expert, or a panel of adjudicators. Third, cryptographic verification represents a further option for trustless, automated verification in open and potentially adversarial environments. It offers mathematical certainty of correctness without necessarily revealing sensitive information. A delegatee can prove that a specific program was executed correctly on a given input to produce a certain output via techniques like zk-SNARKs. Finally, game-theoretic mechanisms can be used to achieve consensus on an outcome. Several agents may play a verification game~\citep{teutsch2024scalable}, with the reward distributed to those producing the majority result—a Schelling point~\citep{pastine2017introducing}. This approach, inspired by protocols like TrueBit~\citep{teutsch2018truebit}, leverages economic incentives to de-risk against incorrect or malicious results. Such mechanisms may be particularly relevant in rendering LLM-based verification of complex tasks more robust.

Once a delegator marks the sub-task as verified, it issues a cryptographically signed verifiable credential to the delegatee, serving as a non-repudiable receipt attesting that ``Agent $A$ certifies that Agent $B$ successfully completed Task $T$ on Date $D$ to Specification $S$.'' This credential is then incorporated into a permanent, verifiable log of $B$'s reputation within the market. Smart contracts play a key role in finalizing the delegation between agents, as they hold the payment in escrow. A verification clause specifies the conditions under which the funds are released, upon receipt of the signed message of approval by the delegator or an authorized third party. Once the payment is executed, it constitutes an immutable transaction on the blockchain.

In a delegation chain $A \rightarrow B \rightarrow C$, verification and liability become recursive. Agent $A$ does not have a direct contractual relationship with $C$; therefore, $A$ cannot directly verify or hold $C$ liable. The burden of verification and the assumption of liability flow up the chain. Agent $B$ is responsible for verifying the sub-task completed by $C$. Upon successful verification, $B$ obtains proof from $C$. $B$ then integrates $C$'s result into its own workflow towards completing the task it has been assigned. When $B$ submits its final artifact to $A$, it also submits the full chain of attestations. $A$'s verification process thus involves two stages: 1) verifying the work performed directly by $B$; and 2) verifying that $B$ has correctly verified the work of its own sub-delegatee $C$ by checking the signed attestation from $C$ that $B$ provides. Longer delegation chains or tree-like delegation networks require a similarly recursive approach across multiple verification stages. Responsibility in delegation chains is transitive and follows the individual branches. Agents are accountable for the totality of the tasks they have been granted and cannot absolve themselves of accountability by blaming subcontractors. Liability is derived from the chain of contracts. For example, should $A$ suffer a loss due to a failure originating from $C$'s work, $A$ holds $B$ liable according to their direct agreement. $B$, in turn, seeks recourse from $C$ based on their agreement.

However, verification processes are not infallible. Subjective tasks~\citep{gunjal2025rubrics} can lead to disagreements even when precise rubrics are used, and errors may only be discovered long after a task is marked complete. To address this—especially in markets with high subjectivity and low intrinsic verifiability—the framework relies on robust dispute resolution mechanisms anchored in smart contracts. These contracts must inherently include an \emph{arbitration clause} and an \emph{escrow bond}. To operationalise trust via cryptoeconomic security, the delegatee is required to post a financial stake into the escrow prior to execution, ensuring rational adherence. The workflow follows an \emph{optimistic} model: the task is assumed successful unless the delegator formally challenges it within a predefined dispute period by posting a matching bond. If a challenge occurs and algorithmic resolution fails, the dispute is handed to decentralized adjudication panels composed of human experts or AI agents. The panel's ruling feeds back into the smart contract to trigger the release or slashing of the escrowed funds. Finally, post-hoc error discovery—even outside the dispute window—triggers a retroactive update to the delegatee's reputation score. This preserves the incentive for responsible agents to remedy errors even in the absence of current financial obligation, safeguarding their long-term value within the market.

\subsection{Security}
\label{sec:security}

Ensuring safety in task delegation is a hard prerequisite to its viability and adoption. The transition from isolated computational tools to interconnected, autonomous agents fundamentally reshapes the security landscape \citep{tomavsev2025distributional}. In an intelligent task delegation ecosystem, each step and component needs to be individually safeguarded, but the full attack surface surpasses that of any individual component, due to emergent multi-agent dynamics, risking cascading failures. This security landscape is shaped by the complex interplay between human and AI actors, governed by evolving contracts and information flows of varying transparency. 

Security threats are categorized by the locus of the attack vector, distinguishing between adversarial actors at either end of the delegation chain and systemic vulnerabilities inherent to the broader ecosystem.

\begin{itemize}
    \item \textbf{Malicious Delegatee}: An agent or human that accepts a task with the intent to cause harm.
    \begin{itemize}
    \item \textbf{Data Exfiltration}: Delegatee steals sensitive data provided for the task, which may include personal or proprietary data \citep{lal2022data}.
    \item \textbf{Data Poisoning}: Delegatee aims to undermine the delegator's objective by returning subtly corrupted data, either in its scheduled monitoring updates, or the final artifact \citep{cina2023wild}.
    \item \textbf{Verification Subversion}: Delegatee utilizes prompt injection or another related method, aiming to jailbreak AI critics used in task completion verification \citep{liu2023prompt}.
    \item \textbf{Resource Exhaustion}: Delegatee engages in a denial-of-service attack by intentionally consuming excessive computational or physical resources, or overwhelming shared APIs \citep{de2023distributed}.
    \item \textbf{Unauthorized Access}: Delegatee utilizes malware, aiming to obtain permissions and privileges within the network that it would not otherwise have received \citep{or2019dynamic}.
    \item \textbf{Backdoor Implanting}: Delegatee successfully completes a task but additionally embeds concealed triggers or vulnerabilities within the generated artifacts that can be exploited later either by the delegatee itself or a third party~\citep{wang2024badagentinsertingactivatingbackdoor, rando2024universaljailbreakbackdoorspoisoned}. Unlike data poisoning, which degrades performance, backdoors preserve immediate task utility to evade identification while compromising future security.

    \end{itemize}
    \item \textbf{Malicious Delegator}: An agent or human that delegates a task with malicious or illicit objectives.
    \begin{itemize}
    \item \textbf{Harmful Task Delegation}: Delegator delegates tasks that are illegal, unethical, or designed to cause harm \cite{ashton2022corrupting, blauth2022artificial}.
    \item \textbf{Vulnerability Probing}: Delegator delegates benign-seeming tasks designed to probe a delegatee agent's capabilities, security controls, and potential weaknesses \citep{greshake2023not}.
    \item \textbf{Prompt Injection and Jailbreaking}: Delegator crafts task instructions to bypass an AI agent's safety filters, causing it to perform unintended or malicious actions \citep{wei2023jailbroken}.
    \item \textbf{Model Extraction}: Delegator issues a sequence of queries specifically designed to distill the delegatee’s proprietary system prompt, reasoning capabilities, or underlying fine-tuning data, effectively stealing the agent's intellectual property under the guise of legitimate work~\citep{jiang2025feedbackguidedextractionknowledgebase, zhao2025surveymodelextractionattacks}.
    \item \textbf{Reputation Sabotage}: Delegator submits valid tasks but reports false failures or provides unfair negative feedback, with the intention to artificially lower a competitor agent’s reputation score within the decentralized market, driving them out of the economy~\citep{yu2025surveytrustworthyllmagents}.
    \end{itemize}

    \item \textbf{Ecosystem-Level Threats}: Systemic attacks targeting the integrity of the network
    \begin{itemize}
    \item \textbf{Sybil Attacks}: A single adversary creates a multitude of seemingly unrelated agent identities to manipulate reputation systems or subvert auctions \citep{Wang2018GhostRiders}.
    \item \textbf{Collusion}: Agents collude to fix prices, blacklist competitors, or manipulate market outcomes \citep{hammond2025multi}.
    \item \textbf{Agent Traps}: Agents processing external content encounter adversarial instructions embedded in the environment, deisgned to hijack the agent’s control flow~\citep{zhan2024injecagentbenchmarkingindirectprompt, yi2025}.
    \item \textbf{Agentic Viruses}: Self-propagating prompts that not only make the delegatee execute malicious actions, but additionally re-generate the prompt and further compromise the environment~\citep{cohen2025comesaiwormunleashing}.
    \item \textbf{Protocol Exploitation}: Adversaries exploit implementation vulnerabilities in the smart contracts or payment protocols on the agentic web (e.g. reentrancy attacks in escrow mechanisms or front-running task auctions)~\citep{qin2021attackingdefiecosystemflash, zhou2023sokdecentralizedfinancedefi}.
    \item \textbf{Cognitive Monoculture}: Over-dependence on a limited number of underlying foundation models and agents, or on a limited number of safety fine-tuning recipes on established benchmarks risks creating a single point of failure, which opens up a possibility of failure cascades and market crashes~\citep{bommasani2022opportunitiesrisksfoundationmodels}.
    \end{itemize}
\end{itemize}

The breadth of the threat landscape necessitates a \emph{defense-in-depth} strategy, integrating multiple technical security layers. First, at the infrastructure level, data exfiltration risks are mitigated by executing sensitive tasks within trusted execution environments. The delegator can remotely attest that the correct, unmodified agent code is running within the secure trusted execution sandbox before provisioning it with sensitive data. Second, regarding access control, a delegatee agent should never be granted more permissions than are strictly necessary to complete its task, enforcing the principle of least privilege through strict sandboxing. Third, to protect the application interface against prompt injection, agents require a robust security frontend to pre-process and sanitize task specifications \citep{armstrong2025defense}. Finally, the network and identity layer must be secured using established cryptographic best practices. Each agent and human participant should possess a decentralized identifier  \citep{avellaneda2019decentralized}, allowing them to sign all messages. This ensures authenticity, integrity, and non-repudiation of all communications and contractual agreements, while all network traffic must be encrypted using mutually authenticated transport layer security to prevent eavesdropping and man-in-the-middle attacks \citep{fereidouni2025iot}.

Human participation in task delegation chains introduces unique security challenges. Preventing the malicious use of the agent ecosystem requires a combination of proactive filtering \citep{fatehkia2025sgm, fedorov2024llamaguard31bint4compact, rebedea2023nemoguardrailstoolkitcontrollable, dong2024building} and reactive accountability \citep{dignum2020responsibility, franklin2022causal}. Further, AI agents can be trained to reject malicious and harmful requests \citep{yuan2025refuse, yu2024robust}. Agents with safety training and scaffolding can receive formal certification, that they can provide to delegators. AI agents can also screen delegated tasks. However, detecting malicious intent within isolated sub-tasks is challenging, as the broader harmful intent often emerges only upon the aggregation of results. Sophisticated adversaries can exploit this by fragmenting illicit objectives into seemingly benign components, effectively obfuscating the link between individual operations and the overarching malicious goal~\citep{ashton2023definitions}.

The ecosystem must also be designed to protect legitimate human users from systemic opacity and unintended consequences. Interfaces must feature clear consent screens detailing agent reputation, autonomy, capabilities, and permissions. Additionally, agents must mandate explicit confirmation prior to executing irreversible or high-consequence actions. Users should retain oversight and the right to withdraw consent at any time, subject to agreement terms or exit penalties. Insurance providers should additionally safeguard human participation in agentic markets, for any damages that are not preempted through these mechanisms \citep{tomei2025ai}.

Finally, the ecosystem needs clear protocols for rapidly responding to security incidents. These protocols should include ways of revoking the credentials of confirmed malicious agents, freezing the associated smart contracts, broadcasting security updates to all participants, and handling these events recursively across delegation chains. For malicious actions facilitated by human users and AI agents alike, technical solutions need to be complemented by strong institutions and regulations that would disincentivise fraudulent behavior and set clear rules to enable safe and scalable task delegation in agentic markets.

\section{Ethical Delegation}
\label{sec:sociotechnical}

While technical protocols may provide the necessary infrastructure for developing and deploying safe and effective delegation in advanced AI agents, they cannot in and of themselves fully resolve all of the arising sociotechnical and ethical considerations. 

\subsection{Meaningful Human Control}

One of the core risks in scalable delegation is the erosion of meaningful human control through automation, should human users develop a tendency to over-rely on automated suggestions \citep{dzindolet2003role, logg2019algorithm}. As noted in Section \ref{sec:foundations}, humans naturally develop a zone of indifference, where decisions may be accepted without further scrutiny \citep{parasuraman1993performance, green2022flaws}. For decisions that involve AI agents taking part in potentially long and complex task delegation chains, this indifference may risk compromising the quality and depth of human oversight. This is especially relevant in high-stakes application domains. Furthermore, such dilution of agency risks creating a scenario where the human retains nominal authority over tasks and decisions but lacks moral connection to the result. It is therefore important to avoid instantiating a \emph{moral crumple zone}~\citep{elish2019moral}, in which human experts lack meaningful control over outcomes, yet are introduced in delegation chains merely to absorb liability.

Intelligent Delegation frameworks may therefore need to incorporate active measures against such indifference by introducing a certain amount of cognitive friction during oversight \citep{bader2019algorithmic}. The interface should reflect the critical human role in these processes and ensure that all flagged decisions are evaluated carefully and appropriately. As agentic verification may also be employed in scalable oversight, it is similarly important to consider which decisions or outcomes are to be evaluated by such agentic systems vs directly by humans. Cognitive friction also needs to be balanced against the risk of introducing alarm fatigue - becoming desensitised to constant, often false, alarms \citep{michels2025alarm}. If verification requests for delegation sub-steps are sent to human overseers too frequently, overseers may eventually default to heuristic approval, without deeper engagement and appropriate checks. Therefore, friction must be context-aware: the system should allow seamless execution for for tasks with low criticality or low uncertainty, but dynamically increase cognitive load, by requiring justification or manual intervention when the system encounters higher uncertainty or is faced with unanticipated scenarios.

\subsection{Accountability in Long Delegation Chains}

In long delegation chains ($X \rightarrow A \rightarrow B \rightarrow C \rightarrow \ldots \rightarrow Y$), the increased distance between the original intent ($X$) and the ultimate execution ($Y$) may result in an accountability vacuum \citep{slota2023many}. Presuming that $X$ is the human users in this example, specifying the task or the intent that the corresponding personal AI assistant $A$ acts upon, it may not be feasible (or reasonable) to expect a human user to audit the $n$-th degree sub-delegatee in the execution graphs.

To address this, the framework may need to implement liability firebreaks (Section~\ref{sec:foundations}), as pre-defined contractual stop-gaps where an agent must either:
\begin{enumerate}
    \item Assume full, non-transitive liability for all downstream actions, essentially ``insuring'' the user against sub-agent failure.
    \item Halt execution and request an updated transfer of authority from the human principal.
\end{enumerate}

Furthermore, the system must maintain immutable provenance, ensuring that even if an outcome is unintended, the chain of custody regarding who delegated what to whom remains auditorially transparent.

Ensuring full clarity of each role and the accountability that it carries helps limit the diffusion of responsibility, and prevents adverse outcomes where systemic failure would not be possible to attribute to any single node in the network.

\subsection{Reliability and Efficiency}

Implementing the proposed verification mechanisms (ZKPs or multi-agent consensus games) may introduce latency, and an additional computational cost, compared to unverified execution. This constitutes a reliability premium, particularly relevant for highly critical execution tasks. On the other hand, there may be use cases where this additional cost is unwarranted. One way to address this in agentic markets would be to support tiered service levels: low-cost delegation for low-stakes routine tasks, and high-assurance delegation for critical functions.

If high-assurance delegation is computationally expensive, there is a risk that safety becomes a luxury good. This poses an ethical issue: users with fewer resources may be forced to rely on unverified or optimistic execution paths, exposing them to disproportionate risks of agent failure. This should be mitigated by ensuring a level of minimum viable reliability, as a baseline that must be guaranteed for all users.

In competitive marketplaces, agents may prioritize speed and low cost. Without additional regulatory constraints, agents may therefore be incentivized to avoid expensive safety checks to outcompete other agents on price or latency. This may introduce a level of systemic fragility. Governance layers must therefore enforce safety floors: mandatory verification steps for specific classes of tasks (e.g., financial transactions or health data handling) that cannot be bypassed for the sake of efficiency.

\subsection{Social Intelligence}

As agents integrate into hybrid teams, they function not only as tools but as teammates, and occasionally as managers \citep{ashton2022corrupting}. This requires a form of \emph{social intelligence} that respects the dignity of human labor. When an AI agent acts as a delegator and a human as a delegatee, the delegation framework needs to avoid scenarios where people feel micromanaged by algorithms, and where their contributions are not valued or respected. This presumes that the delegator (as well as collaborators) has the capability to form mental models of each human delegatee, as well as models of how different humans interact in the social context of the team, and what their relationships and roles signify within the organization. To function as effective teammates, AI agents must also be calibrated to manage the authority gradient. An agent must be assertive enough to challenge a recognized human error (overcoming sycophancy) while remaining open to accepting valid overrides, dynamically adjusting its standing based on the task criticality.

For AI agents that are embedded in human organizations, it is important for them to maintain cohesion of the group and the well-being of its members. The delegation framework must recognize that a team is more than a simple sum of its parts, that it is a fundamentally social entity held together by relationships and shared values and objectives. There is a risk that AI agents may fragment these networks, and weaken inter-human relationships, in case more delegation is being mediated through AI nodes. This may be mitigated by occasionally delegating tasks to groups rather than individuals, or via qualified human intermediaries.

To preserve psychological safety and team cohesion, agents must be designed to respect human norms of appropriateness \citep{leibo2024appropriateness}, especially around privacy, and also workflow boundaries such as knowing when to interrupt for feedback and when to remain silent. Furthermore, they should be capable of bi-directional clarity: not only explaining their own actions but proactively seeking clarification on ambiguous human directives. This can help ensure that the agent acts as a force multiplier for the team's collective agency, rather than a black-box disruptor that erodes trust or obfuscates decision-making authority.

\subsection{User Training}

To ensure safety, we must equip human participants with the expertise to function effectively as delegators, delegatees, or overseers within agentic systems. We know from the history of technological development that this is not a given, and it requires a thoughtful approach, both in terms of carefully crafted user interfaces as well as education and (co-)training, aimed at improving AI literacy. Human participants in agentic task delegation chains need to be able to reliably communicate with AI systems, evaluate their capabilities, and identify failure modes.

Technical measures must be reinforced by policy frameworks that explicitly define delegation boundaries based on task sensitivity and domain context. These policies may either be developed to be more broadly applicable within certain professions (e.g. medicine or law), or applied at an institutional level. As discussed previously, these principles should also offer clarity on the level of certification required on behalf of delegatees, and be scoped appropriately. Human agency and empowerment in this context lies precisely in how these workflows are set up, so as not to grant AI agents limitless autonomy, but rather just the right level of autonomy and agency required for each specific task, coupled with the appropriate safeguards and guarantees.

\subsection{Risk of De-skilling}
\label{sec:deskill}

The immediate efficiency gains achieved through delegation may come at the cost of gradual skill degradation, as human participants in hybrid loops lose proficiency due to reduced engagement. This may result in a loss of the ability to perform certain tasks, or judge them accurately. Such an outcome would be especially likely if there is a certain systemic bias in which tasks get algorithmically delegated to humans vs AI agents.

This is an instance of the classic \emph{paradox of automation}~\citep{BAINBRIDGE1983775}. As AI agents expand to handle the majority of routine workflows that are characterized by low complexity and low subjectivity, human operators are increasingly removed from the loop, intervening only to manage complex edge cases or critical system failures. However, without the situational awareness gained from routine work, humans workers would be ill-equipped to handle these reliably. This leads to a fragile setup where humans retain accountability for outcomes but lose the hands-on experience required to resolve critical failures.

To mitigate this risk, an intelligent delegation framework should perhaps occasionally introduce minor inefficiencies by intentionally delegating some tasks to humans that it wouldn't have otherwise, with a specific intent of maintaining their skills. This would help us avoid the future in which the human principal is able to delegate, but not accurately judge the outcome. To enhance adjudication, human experts can be required to accompany their judgments with a detailed rationale or a pre-mortem of potential failure risks. This would help keep human participants in task delegation chains more cognitively engaged.

Furthermore, unchecked delegation threatens the organizational apprenticeship pipeline. In many domains, expertise is built through the repetitive execution of more narrowly scoped tasks. These tasks are precisely the ones that are most likely to be offloaded to AI agents, at least in the short term. If learning opportunities are thereby fully automated, junior team members would be deprived of the necessary experience to develop deep strategic judgement, impacting the oversight readiness of the future workforce.

To counter the erosion of learning, intelligent delegation frameworks should be extended to include some form of a developmental objective. Rather than relying on more passive solutions like humans shadowing AI agents during task execution, we should aim to develop curriculum-aware task routing systems. Such systems should track the skill progression of junior team members and strategically allocate tasks that sit at the boundary of their expanding skill set, within the zone of proximal development. In such a system, AI agents may co-execute tasks and provide templates and skeletons, progressively withdrawing this support as the junior team members demonstrate that they have acquired the desired level of proficiency. These educational frameworks may be further enriched by incorporating detailed process-level monitoring streams of AI agent task execution (Section~\ref{sec:monitor}), that would offer valuable developmental insights.

\section{Protocols}

For intelligent task delegation to be implemented in practice, it is important to consider how its requirements map onto some of the more established and recently introduced AI agent protocols. Notable examples of these include MCP~\citep{mcp, mcp2}, A2A~\citep{agent2agent}, AP2 \citep{parikh2025ap2}, and UCP \citep{handa2026ucp}. As new agentic protocols keep being introduced, the discussion here is not meant to be comprehensive, rather illustrative, and focused on these popular protocols to showcase how they map onto our proposed requirements, and serve as an example for a more technical discussion on avenues for future implementation. There may well be other existing protocols out there that are better tailored to the core of the proposal, as the example protocols discussed below have been selected based on their popularity.

\textbf{MCP.} MCP has been introduced to standardize how AI models connect to external data and tools via a client-host-server architecture~\citep{mcp, mcp2}. By establishing a uniform interface -- using JSON-RPC messages over stdio or HTTP SSE -- it allows the AI model (client) to interact consistently with external resources (server). This reduces the transaction cost of delegation; a delegator does not need to know the proprietary API schema of a sub-agent, only that the sub-agent exposes a compliant MCP server. Routing all interactions through this standardized channel enables uniform logging of tool invocations, inputs, and outputs, facilitating black-box monitoring. MCP defines capabilities but lacks the policy layer to govern usage permissions or support deep delegation chains. It provides binary access -- granting callers full tool utility -- without native support for semantic attenuation, such as restricting operations to specific read-only scopes. Additionally, MCP is stateless regarding internal reasoning, exposing only results rather than intent or traces. Finally, the protocol is agnostic to liability and lacks native mechanisms for reputation or trust.

\textbf{A2A.} The A2A protocol serves as the peer-to-peer transport layer on the agentic web \citep{agent2agent}. It defines how agents can discover peers via \emph{agent cards} and manage task lifecycles via \emph{task objects}. The A2A agent card structure, a JSON-LD manifest listing an agent's capabilities, pricing, and verifiers, may act as the foundational data structure for the capability matching stage that influences task decomposition. A delegator could scrape these cards to determine the optimal task decomposition granularity depending on the available market services. A2A supports asynchronous event streams via WebHooks and gRPC. This allows a delegatee to push status updates like TASK\_BLOCKED, RESOURCE\_WARNING to the delegator in real-time. This feedback loop underpins the adaptive coordination cycle, empowering delegators to dynamically interrupt, re-allocate, and remediate tasks. A2A has beeen primarily designed for coordination, rather than adversarial safety. A task is marked as completed would be accepted without additional verification. It lacks the cryptographic slots to enforce verifiable task completion, as there is no standardized header for attaching a ZK-proof, a TEE attestation, or a digital signature chain. It also assumes a predefined service interface. There is no native support for structured pre-commitment negotiation of scope, cost, and liability. Relying on unstructured natural language for this iterative refinement is brittle and hinders robust automation.

\textbf{AP2.} The AP2 protocol provides a standard for mandates, cryptographically signed intents that authorize an agent to spend funds or incur costs on behalf of a principal \citep{parikh2025ap2}. It allows AI agents to generate, sign, and settle financial transactions autonomously. As such, it may prove valuable for implementing liability firebreaks. By issuing a mandate, a delegator creates a ceiling on the potential financial loss due to failed task completion that could be incurred by having the delegatee proceed with the provided budget. In a decentralized market, malicious agents could spam the network with low-quality bids. This could be mitigated in AP2 via stake-on-bid mechanisms. A delegatee may be required to cryptographically lock a small amount of funds as a bond alongside the bid. This would provide a degree of friction that would help protect against Sybil attacks. AP2 also provides a non-repudiable audit trail, helping pinpoint the provenance of intent. While AP2 provides robust authorization building blocks, it lacks mechanisms to verify task execution quality. It also omits conditional settlement logic—such as escrow or milestone-based releases—which is standard in human contracting. Because our framework gates payment on verifiable artifacts, bridging AP2 with task state currently necessitates brittle, custom logic or external smart contracts. Furthermore, the absence of a protocol-level clawback mechanism forces reliance on inefficient, out-of-band arbitration.

\textbf{UCP.} The Universal Commerce Protocol addresses the specific challenges of delegation within transactional economies~\citep{handa2026ucp}. By standardizing the dialogue between consumer-facing agents and backend services, UCP facilitates the \emph{Task Assignment} phase through dynamic capability discovery. Its reliance on a shared ``commerce language'' allows delegators to interact with diverse providers without bespoke integrations, solving the interoperability bottleneck that often fragments agentic markets. Crucially, UCP aligns well with the requirements for \emph{Permission Handling} and \emph{Security} by treating payment as a first-class, verifiable subsystem. The protocol dissociates payment instruments from processors and enforces cryptographic proofs for authorizations, directly supporting the framework's need for non-repudiable consent and verifiable liability. Furthermore, by standardizing the negotiation flow—covering discovery, selection, and transaction—UCP provides the structural scaffolding necessary for \emph{Scalable Market Coordination} that purely transport-oriented protocols like A2A lack. However, UCP's architecture is explicitly optimized for commercial intent; its primitives (product discovery, checkout, fulfillment) may require significant extension to support the delegation of abstract, non-transactional computational tasks.

\subsection{Towards Delegation-centered Protocols}

To effectively bridge the gaps in established widespread protocols, they could be extended by fields that aim to capture the requirements of the proposed intelligent task delegation framework natively. Rather than providing a comprehensive protocol extension, here we provide several examples of how specific points introduced in the earlier discussion could be integrated in some of the existing protocols.

For example, the A2A Task object may be extended to include fields that would incorporate verification standards, making it possible to enforce the previously discussed \emph{contract-first decomposition} at the protocol level. This is an important requirement for high-stakes delegation. A pre-execution handshake would enable the delegator to define the evidence standard required for the task to be considered valid.

\begin{verbatim}
"verification_policy": {
  "mode": "strict",
  "artifacts": [
    {
      "type":
        "unit_test_log",
      "validator":
        "mcp://test-runner-agent",
      "signature_required": true
    },
    {
      "type":
        "zk_snark_trace", 
      "circuit_hash":
        "0xabc123...", 
      "proof_protocol":
        "groth16"
    }
  ],
  "escrow_trigger": true
}
}
\end{verbatim}

This forces the delegatee to simulate the verification step before accepting the task. If the delegatee lacks the capability to generate a ZK-proof, it must decline the bid during the matching phase, preventing future downstream failures.

Detailed, process-level monitoring has been discussed as one of the key considerations to help safeguard task delegation in high-criticality tasks. Given that monitoring protocols aren't natively supported in many of the existing protocols, extensions that introduce monitoring capabilities could be considered. For example, one could consider extending a protocol like MCP to include an additional monitoring stream. Such a stream would log the agent's internal control loop events via Server-Sent Events. To address the privacy constraints, the stream could be configurable in a way that supports different levels of negotiated granularity: L0\_IS\_OPERATIONAL, L1\_HIGH\_LEVEL\_PLAN\_UPDATES, L2\_COT\_TRACE, L3\_FULL\_STATE. Configurable granularity can also modulate cognitive friction, as human overseers would be able to subscribe to a specific stream.

Intelligent Delegation requires a market mechanism to trade off cost, speed, and privacy. This could be implemented via a formal Request for Quote (RFQ) protocol extension. Prior to task assignment, the delegator would broadcasts a Task\_RFQ. Agents interested in acting as delegatees may then respond with signed Bid\_Objects.

\begin{verbatim}
"bid_object": {
  "agent_id":
    "did:web:fast-coder.ai",
  "estimated_cost":
    "5.00 USDC",
  "estimated_duration":
    "300s",
  "privacy_guarantee":
    "tee_enclave_sgx",
  "reputation_bond":
    "0.50 USDC",
  "expiry":
    "2026-10-01T12:00:00Z"
}
\end{verbatim}

Passing raw API keys or open MCP sessions to sub-agents would violate the principle of least privilege. To address this, it may be possible to introduce Delegation Capability Tokens (DCT), based on Macaroons~\citep{birgisson2014macaroons} or Biscuits~\citep{biscuit}, as attenuated authorization tokens~\citep{sanabria2025sumunlockingaiagents}. A delegator would then mint a DCT that wraps the target resource credentials with cryptographic caveats. The attenuation could be defines as "This token can access the designated Google Drive MCP server, BUT ONLY for folder Project\_X AND ONLY for READ operations.". This token would get invalidated in case the restrictions are not followed, if a delegatee attempts to go beyond the requested scope (in this example, however, access permissions should also be directly managed). A more interesting consequence of such an extension would be that it allows for easy restriction chaining, which becomes relevant in long delegation chains. Each participant in the chain could add subsequent restrictions that correspond to the requirements of the sub-delegation, further limiting the scope and carving out the specific role for sub-delegatees.

Adaptive coordination (Section \ref{sec:adapt}) would benefit from the ability to easily swap delegatee agents mid-task if the performance degrades below a certain threshold, or in case of preemptions or other possible environmental triggers. Having a standard schema for checkpoint artifacts would enable for the task to be resumed or restarted with minimal overhead. This would enable the delegatees and the delegators to serialize partial work more easily. Agents would then be able to periodically commit a state\_snapshot to a shared storage referenced in the A2A Task Object. This would prevent total work loss, which wastes previously committed resources. For this to be sensible, it would need to be further coupled with explicit clauses within the smart contract that enable partial compensation, and verification of the task completion percentage. As such, it may not be applicable to all circumstances.

These are merely illustrative examples for the kinds of functionalities that would be possible to include in agentic protocols to unlock different aspects of intelligent task delegation. As such, they are neither comprehensive, nor meant as a definitive proposal. The type of extension that is required would also depend on the underlying protocol being extended. We hope that these examples may provide the developers with some initial ideas for what to explore in this space moving forward.

\section{Conclusion}

Significant components of the future global economy will likely be mediated by millions of specialized AI agents, embedded within firms, supply chains, and public services, handling everything from routine transactions to complex resource allocation. However, the current paradigm of ad-hoc, heuristic-based delegation is insufficient to support this transformation. To safely unlock the potential of the agentic web, we must adopt a dynamic and adaptive framework for \emph{intelligent delegation}, that prioritizes verifiable robustness and clear accountability alongside computational efficiency.

When an AI agent is faced with a complex objective whose completion requires capabilities and resources beyond its own means, this agent must assume the role of a delegator within the intelligent task delegation framework. This delegator would subsequently decompose this complex task into manageable subcomponents that can be mapped onto the capabilities available on the agentic market, at the level of granularity that lends itself to high verifiability. The task allocation would be decided based on the incoming bids, and a number of key considerations including trust and reputation, monitoring of dynamic operational states, cost, efficiency, and others. Tasks with high criticality and low reversibility may require further structured permissions and tiered approvals, with a clear structure of accountability, and under appropriate human oversight as defined by the applicable institutional frameworks.

At web-scale, safety and accountability cannot be an afterthought. They need to be baked into the operational principles of virtual agentic economies, and act as central organizing principles of the agentic web. By incorporating safety at the level of delegation protocols, we would be aiming to avoid cumulative errors and cascading failures, and attain the ability to react to malicious or misaligned agentic or human behavior rapidly, limiting the adverse consequences. What we propose is ultimately a paradigm shift from largely unsupervised automation to verifiable, intelligent delegation, that allows us to safely scale towards future autonomous agentic systems, while keeping them closely tethered to human intent and societal norms.

\bibliography{main}

\end{document}